# PSSF: Early osteoarthritis detection using physical synthetic knee X-ray scans and AI radiomics models

Abbas Alzubaidi[a] and Ali Al-Bayaty[b]

[a] Department of Radiologic Science and Technology, American University of Iraq–Baghdad, Baghdad, Iraq, abbas.alzubaidi@auib.edu.iq
[b] Department of Electrical and Computer Engineering, Portland State University, Portland, OR 97201, USA, albayaty@pdx.edu

**Abstract**

Knee osteoarthritis (OA) is a major cause of disability worldwide and is still largely assessed using subjective radiographic grading, most commonly the Kellgren–Lawrence (KL) scale. Artificial intelligence (AI) and radiomics offer quantitative tools for OA assessment but depend on large, well-annotated image datasets, mainly X-ray scans, that are often difficult to obtain because of privacy, governance and resourcing constraints. In this research, we introduce a physics-based synthetic simulation framework (PSSF) to fully generate controllable X-ray scans without patients' involvement and violating their privacy and institutional constraints. This PSSF is a 2D X-ray projection simulator of anteroposterior knee radiographs from a parametric anatomical model of the distal femur and proximal tibia. Using PSSF, we create a virtual cohort of 180 subjects (260 knees), each is imaged under three protocols (reference, low-dose, and geometry-shift). Medial joint regions are automatically localized, preprocessed, and processed with the Image Biomarker Standardisation Initiative (IBSI). Practically, three machine learning (ML) models are utilized, logistic regression, random forest, and gradient boosting, to train binary (KL-like "0" vs. "2") and three-class (0–2) prediction radiographic images. Robustness is assessed within IBSI protocol, cross-protocol, and multi-protocol scenarios. Finally, features stability is then evaluated using intraclass correlation coefficients across acquisition changes.

## 1. Introduction

Knee osteoarthritis (OA) is a leading global cause of pain, functional limitation, and disability, with prevalence expected to rise substantially in coming decades due to population ageing and increasing obesity rates [1,2]. Radiographic assessment remains central to diagnosis and monitoring, with Kellgren–Lawrence (KL) classification widely used in clinical practice and research [3,4]. KL grading is convenient and clinically familiar, but it is semi-quantitative, reader-dependent, and only moderately reliable in routine use, particularly in the early disease stages where subtle joint space narrowing and small osteophytes are difficult to grade consistently [5,6].

Artificial intelligence (AI) and radiomics offer quantitative tools to complement radiographic OA assessment by extracting reproducible image biomarkers from standard-of-care imaging [7–9]. Radiomics converts images into high-dimensional descriptors of intensity, texture, and morphology that can be associated with disease phenotype and outcome, while deep learning models can learn predictive representations directly from pixels [7–9]. On large cohorts of knee radiographs, convolutional neural networks (CNNs) have demonstrated strong performance for automated KL grading and OA progression prediction, often approaching expert-level agreement [10–15]. Standardisation initiatives such as the Im- age Biomarker Standardisation Initiative (IBSI) and validated open-source implementations have further improved transparency and reproducibility of radiomic pipelines [16–18].

However, radiomics and deep learning pipelines can be sensitive to acquisition and reconstruction variability, leading to degraded reproducibility and cross-site generalisation if protocol effects are not controlled or modelled [19–23]. Systematic reviews and phantom studies have shown that a non-trivial fraction of radiomic features can be unstable under changes in dose, reconstruction kernel, or scanner platform, motivating careful pre-processing, feature stability screening, and harmonisation strategies in multi-centre research [19–22]. Similarly, deep learning models can exploit confounding information and suffer performance drops when evaluated on external institutions or shifted acquisition conditions, underscoring the importance of robustness testing beyond within-site splits [23]. Recent reporting guidance, including TRIPOD, TRIPOD+AI, and CLAIM, also emphasises transparent documentation of data, modelling, and validation choices for clinical prediction modelling and medical imaging AI studies [24–26].

Synthetic medical imaging has been proposed as one way to mitigate data scarcity and privacy limitations in imaging AI and radiomics research. Data-driven generative models, especially generative adversarial networks (GANs) and image-to-image translation approaches such as CycleGAN, can create realistic synthetic images for augmentation and method development [27–29]. In knee OA, Prezja et al. generated "DeepFake" OA knee radiographs using GANs and demonstrated both expert deception and augmentation potential, while later work explored CycleGAN-based synthesis of hypothetical OA states and evaluated augmentation strategies in radiograph classification [30–32]. Nevertheless, purely data-driven synthesis typically still requires access to real radiographs for training and offers limited direct control over physical acquisition parameters and disease morphology.

Physics-based simulation provides a complementary approach by explicitly modelling the anatomy and the radiographic imaging chain, enabling deterministic manipulation of disease morphology and acquisition conditions with known ground-truth labels. Such simulators are particularly useful for early-stage methods development, robustness testing, and teaching, and can support systematic evaluation of protocol sensitivity and feature stability before clinical calibration and validation.

In this research, we present a physics-based synthetic simulation framework (PSSF) for generating a fully synthetic cohort of knee radiographs representing early OA (KL-like grades "0–2"), and use this cohort to train and evaluate radiomics-based machine-learning models. Note that a cohort is a patch of training and testing medical datasets. For PSSF, our research objectives are listed as follows.

1. Developing a parametric anatomical and imaging model of the knee suitable for synthetic radiograph generation.

2. Constructing a virtual cohort with controlled early OA morphology and multiple acquisition protocols.

3. Extracting IBSI-compliant radiomic features and train classifiers for early OA grading.

4. Quantifying radiomics features stability and model robustness under acquisition variability.

Because all medical images of PSSF are entirely synthetic, no patients' consents or IRB approvals are required, and these generated data can be shared openly for methodological development and experimental benchmarking.

## 2. Methods

This study is entirely designed using synthetic medical imaging datasets and consists of five main stages, as illustrated in Figure 1. These stages include: (i) definition of a parametric anatomical model of the distal femur, proximal tibia, and joint space, (ii) implementation of a physics-based X-ray projection simulator to generate 16-bit knee radiographs (Figure 2), (iii) generation of a virtual cohort spanning KL-like grades "0–2" under three acquisition protocols (Table 1; Figure 3), (iv) automatic localisation of the medial compartment, IBSI-aligned preprocessing, and radiomic feature extraction (Figure 4) [16–18], and (v) training and evaluation of machine learning models (logistic regression, random forest, and gradient boosting) for early OA classification and robustness assessment (Figure 5; Figure 6) [33–35].

**Table 1.** Virtual synthetic cohort composition and acquisition protocols.

| KL-like grade | Number of knees | Radiographs per knee (protocols) | Total radiographs |
|---|---|---|---|
| 0 (normal) | 130 | 3 (reference, low-dose, geometry shift) | 390 |
| 1 (doubtful OA) | 78 | 3 | 234 |
| 2 (mild OA) | 52 | 3 | 156 |
| Total | 260 | – | 780 |

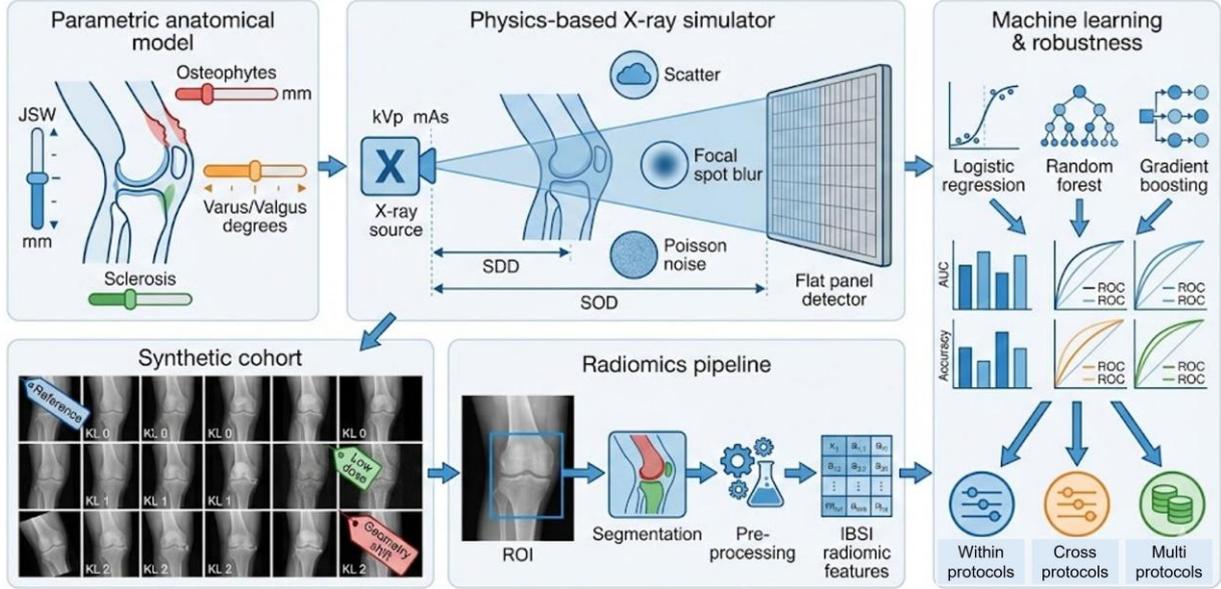

**Figure 1.** The conceptual overview of our physics-based synthetic simulation framework (PSSF). From left- to-right: parametric anatomical model of the knee with adjustable morphology sliders (joint space width, osteophytes, sclerosis, and varus/valgus), physics-based X-ray simulator with tube, knee and flat-panel detector, synthetic cohort of graded knee radiographs under multiple protocols, radiomics pipeline (ROI localization, preprocessing, and IBSI-compliant features extraction), and machine learning (ML) models for early OA classification and robustness experiments.

## 2.1. Parametric knee anatomy and KL-like mapping

### 2.1.1. Anatomical model

The knee joint is represented in 2D as an anteroposterior-like projection of the distal femur and proximal tibia, including femoral condyles, tibial plateau and subchondral bone plates. The joint space is modelled as an explicit gap between opposing bone contours, serving as a surrogate for cartilage thickness. Small bony protrusions at characteristic sites (medial and lateral femoral condyles, tibial spines) represent osteophytes.

A set of continuous parameters controls morphology: medial joint space width "JSW_med", lateral joint space width "JSW_lat", osteophyte count and maximal size, subchondral plate thickness/density "sclerosis proxy", overall varus/valgus angle, and basic shape descriptors of femoral condyles and tibial plateau. Parameter ranges are chosen to be qualitatively consistent with KL descriptors, where grade "0" corresponds to normal joint space and no osteophytes, grade "1" to minimal narrowing or small osteophytes, and grade "2" to definite osteophytes and clear medial joint space narrowing without gross deformity [3,4]. Illustrative parameter ranges used to configure the simulator are provided in Table 2.

*Relative units; higher values indicate thicker, denser subchondral bone.*

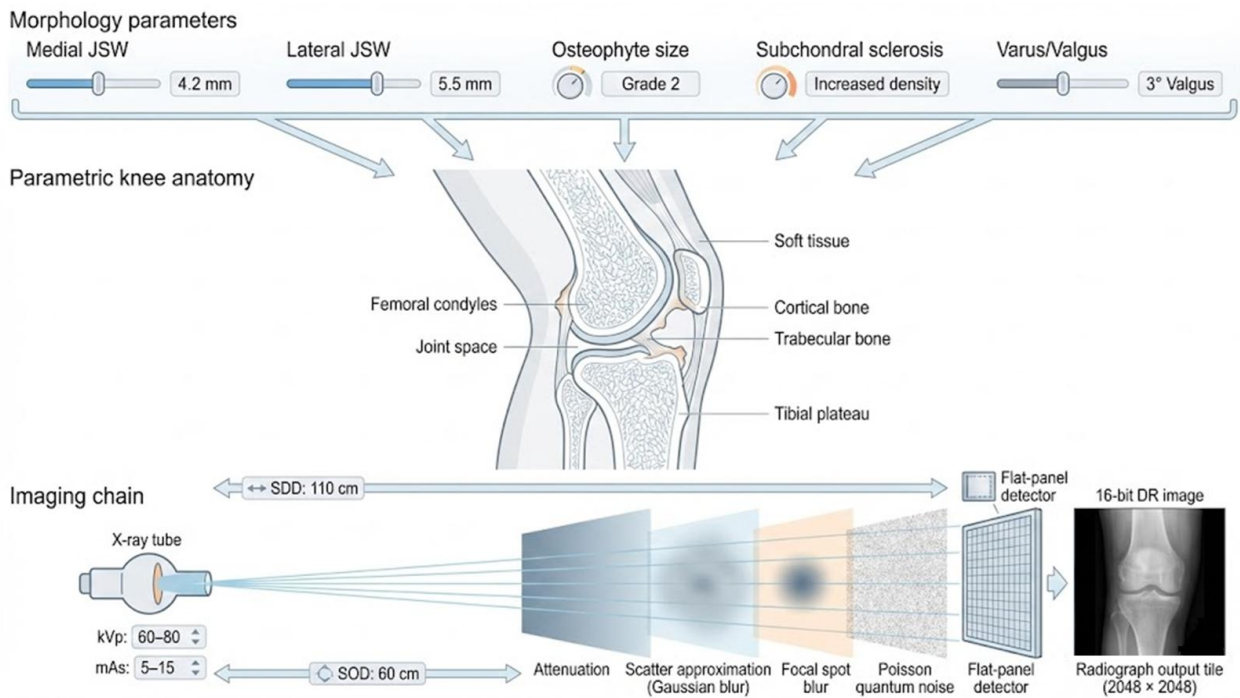

**Figure 2.** The anatomy and imaging chain as three-tier schematics of the PSSF. Top tier: adjustable morphological parameters (medial/lateral JSW, osteophyte size, subchondral sclerosis, and alignment). Middle tier: parametric line-art representation of distal femur, proximal tibia and joint space with labelled cortical bone, trabecular bone and soft tissue. Bottom tier: imaging chain with X-ray source, beam paths through the knee, scatter and blur approximations, Poisson noise and digital flat-panel detector producing a 16-bit radiograph.

**Table 2.** Example parameter ranges used to map KL-like grades to simulate knee morphology.

| Parameter | Grade 0 (normal) | Grade 1 (doubtful OA) | Grade 2 (mild OA) |
| --- | --- | --- | --- |
| JSW_med (mm) | 4.5 – 5.5 | 3.5 – 5.0 | 2.5 – 4.0 |
| JSW_lat (mm) | 4.5 – 5.5 | 4.0 – 5.5 | 3.5 – 5.0 |
| Max osteophyte size (mm) | 0 (absent) | 0 – 1.5 | 1 – 3 |
| Osteophyte count | 0 | 0 – 2 | 1 – 4 |
| Subchondral plate thickness* | baseline (1.0) | 1.0 – 1.3 | 1.2 – 1.6 |
| Varus/valgus angle (degrees) | −1 to +1 | −3 to +3 | −5 to +5 |

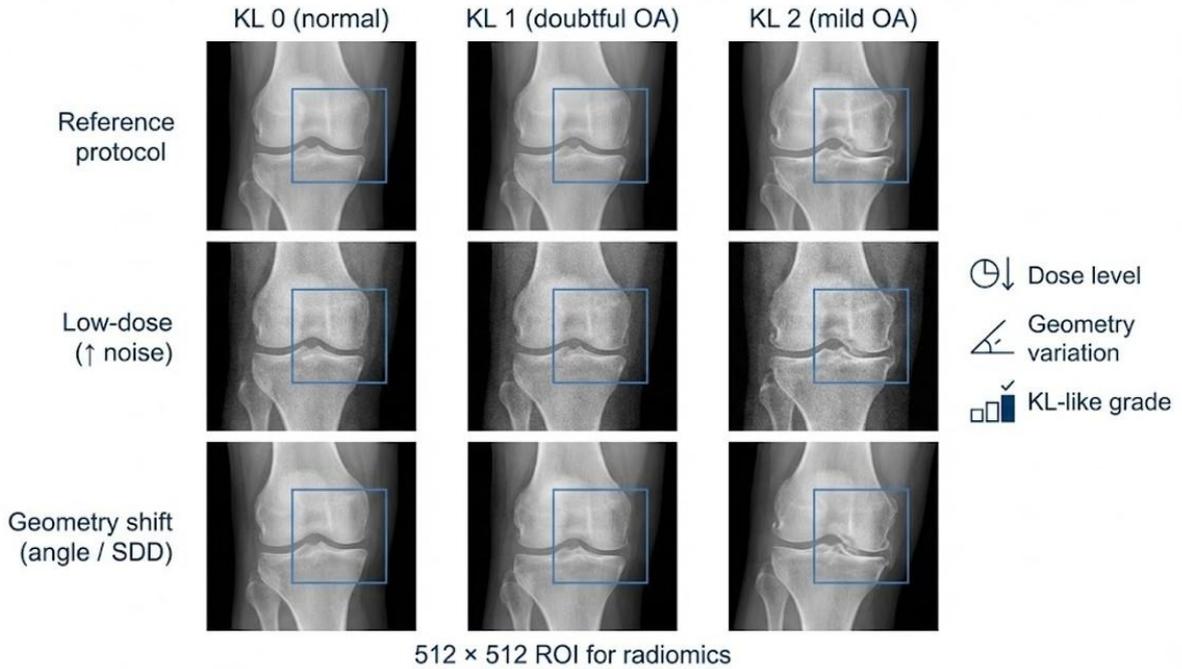

**Figure 3.** Synthetic cohort examples: KL-like grades and acquisition protocols. 3×3 panel of synthetic knee radiographs. Columns represent KL-like grades "0" (normal), "1" (doubtful OA), and "2" (mild OA), where rows represent acquisition protocols (reference, low-dose, and geometry shift). Each cell shows an AP-like knee radiograph with class-appropriate joint space and osteophytes. A colored square ROI indicates the 512×512 medial compartment region used for radiomics.

*2.1.2. Imaging chain*

The imaging chain models a simplified digital radiography system (see Figure 2). An effective poly-energetic X-ray spectrum (60–80 kVp) is combined with material-specific linear attenuation coefficients for cortical bone, trabecular bone and soft tissue. For each detector pixel, transmitted intensity is computed as stated in Eq. (1), where $I_0$ is the initial value of a projective intensity, $t_i(x,y)$ denotes the path length through material $i$ at pixel (x, y), and $\mu_i$ is the corresponding effective attenuation coefficient.

$$I(x, y) = I_0\, e^{-\sum_{i=1} \mu_i\, t_i(x, y)} \tag{1}$$

Note that acquisition parameters include field of view (FOV) 2048×2048 pixels, pixel size of 0.15 mm source-to-detector distance (SDD) 110–120 cm, source-to-object distance (SOD) 90–100 cm, and relative mAs for dose modulation. Scatter is approximated by adding a low-frequency blurred component derived from the primary image. Focal spot blur and detector modulation transfer function are modelled by convolution with a mild spatial kernel. Quantum noise is introduced as Poisson noise applied to transmitted photon counts, followed by additive Gaussian readout noise. Images are stored as 16-bit grayscale arrays and rescaled to a standard dynamic range for analysis.

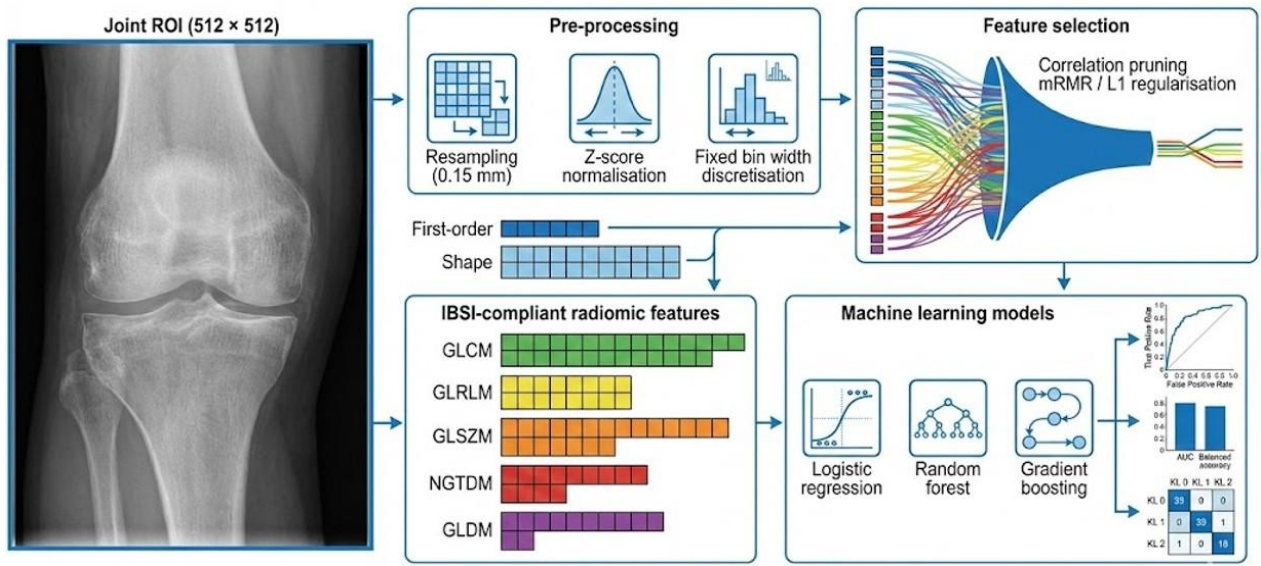

**Figure 4.** Radiomics and ML pipeline. Horizontal infographic showing the radiomics and AI workflow. A magnified medial joint ROI feeds into preprocessing steps (resampling, z-score normalization, and fixed-bin discretization), then into an IBSI-compliant feature matrix organized by a family (first-order, shape, GLCM, GLRLM, GLSZM, NGTDM, or GLDM). A feature-selection funnel reduces dimensionality before feeding logistic regression, random forest and gradient boosting models. Output plots include ROC curves, performance metrics, and three-class confusion matrices.

## 2.2. Synthetic cohort generation

A virtual population of 180 subjects (260 knees) is defined, with KL-like grades randomly assigned so that early disease stages are emphasised. For instance, approximately 50% of knees are KL-like "0" (normal), 30% grade "1" (doubtful OA), and 20% grade "2" (mild OA). For each knee, anatomical parameters are sampled from the grade-dependent ranges, as expressed in Table 2.

Three acquisition protocols are simulated as follows.
1. A reference protocol with nominal kVp, mAs, SDD/SOD, and beam angle.
2. A low-dose protocol with reduced mAs, higher relative noise, and identical geometry.
3. A geometry-shift protocol with a small change in beam angle and SDD, to mimic positioning variability.

Each knee is imaged once under each protocol, yielding 780 radiographs in total. Cohort composition and protocol counts are summarised in Table 1, and representative images are demonstrated in Figure 3.

## 2.3. KL-like grade-morphology mapping

To maintain a consistent link between KL-like grade and simulated anatomy, parameter ranges are defined in Table 2. These can be refined based on expert radiology input or atlas data but serve as a practical template.

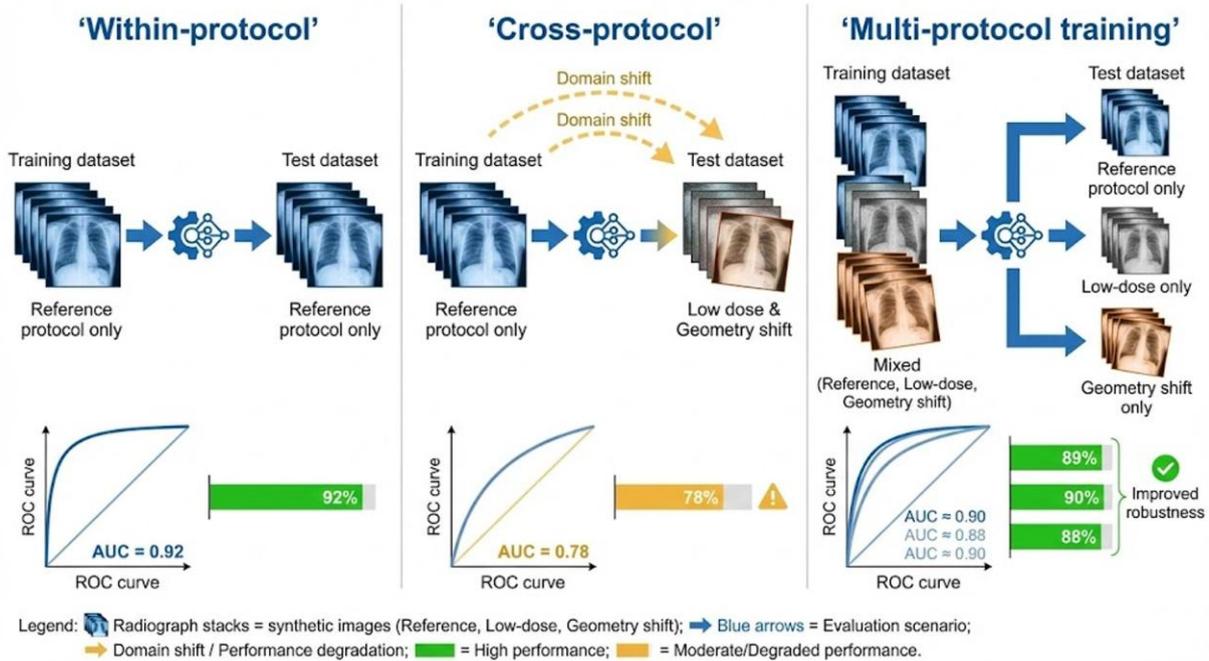

**Figure 5.** Robustness experiments across acquisition protocols. Schematic of three evaluation scenarios. Left panel: within-protocol setting where reference-protocol training and testing achieve high performance. Middle panel: cross-protocol setting where a model trained on reference images is tested on low-dose and geometry-shift images, showing degraded ROC and performance. Right panel: multi-protocol training where models trained on a mixture of all protocols recover improved performance across all test sets, illustrating enhanced robustness to domain shift.

*2.4. ROI localization, preprocessing, and radiomics extraction*

Given known geometry, the medial compartment can be deterministically localised. To emulate a more realistic workflow, an automatic localisation step performs template matching to approximate the joint centre, followed by extraction of a 512×512 pixel square ROI (Region of Interest), which is centred on the medial joint line as shown in Figure 3 and Figure 4.

Preprocessing follows IBSI-aligned practice. Because the simulator produces isotropic pixel spacing, ROIs are analysed at native resolution, intensities within each ROI are standardised (z-score normalisation), and fixed-bin discretisation is applied prior to texture matrix computation [16–18]. Radiomic features are extracted using an IBSI-compliant library and include first-order statistics, 2D shape descriptors, and texture families derived from co-occurrence and run-length based matrices (e.g., GLCM, GLRLM, GLSZM, NGTDM, GLDM) [16–18]. To improve robustness and reduce redundancy, highly correlated features (e.g., $|\rho| > 0.9$) are pruned and the remaining features are standardised across the dataset. When acquisition variability is explicitly studied (low-dose and geometry shift), feature stability screening and harmonisation concepts from the broader radiomics literature motivate the identification of stable feature subsets before model fitting [19–22].

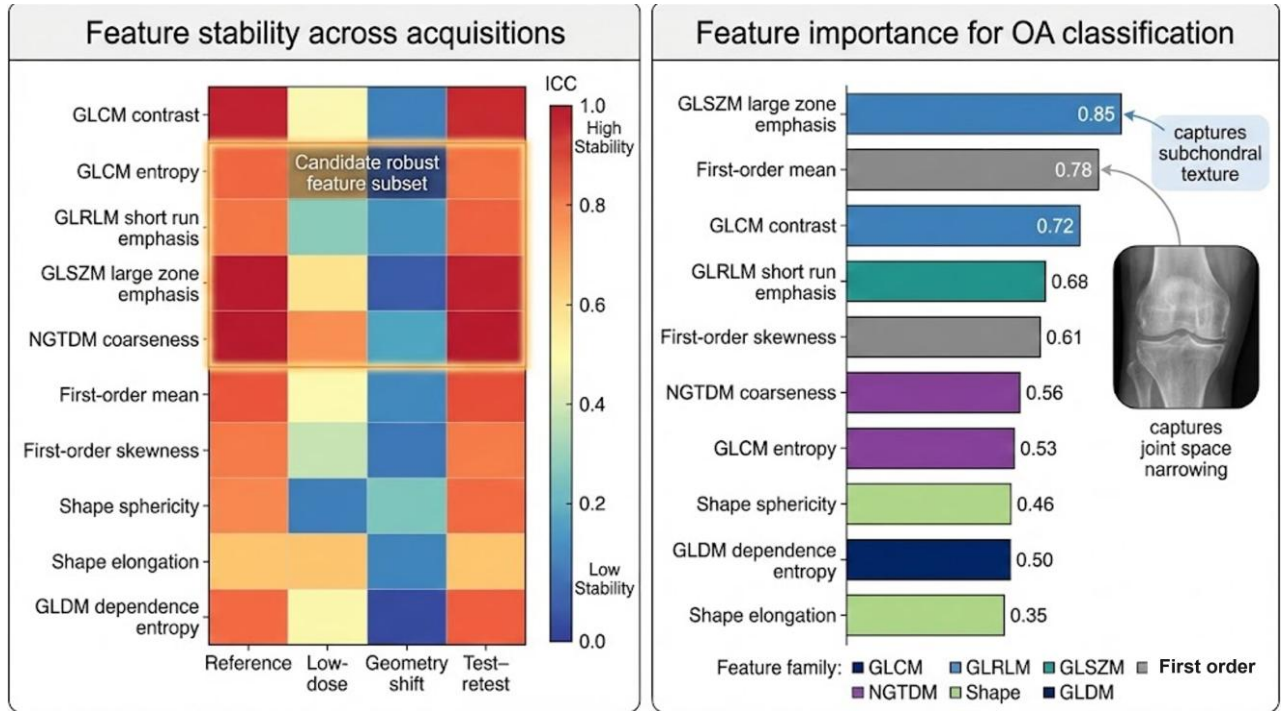

**Figure 6.** Radiomic feature stability and importance. Summary of feature-level behavior. Left panel: heatmap of intraclass correlation coefficients for radiomic features across reference, low-dose, geometry-shift, and test–retest conditions, highlighting a subset of highly stable features. Right panel: bar chart of the most important features for KL-like "0" vs. "2" classification, grouped by feature family. Example features are annotated with their qualitative interpretation, e.g., capturing joint space narrowing or subchondral texture coarseness.

## 2.5. Machine learning models and evaluation

In this research, two prediction tasks are considered: (i) binary classification of KL-like grade "0" vs. "2", and (ii) three-class classification of grades "0", "1", and "2". Three model families are evaluated: logistic regression with L2 regularisation, random forest, and gradient boosting (including XGBoost-style boosted trees) [33–35]. Related supervised ML classifiers and neural-network models have also been applied to imaging feature detection and classification workflows across disease domains, supporting the generality of these modelling choices [36]. Feature selection is performed using L1-regularised logistic regression and minimum redundancy–maximum relevance (mRMR) to yield compact, informative signatures [37]. Models are trained and evaluated using subject-level splitting (70% training, 15% validation, 15% testing) to avoid leakage across knees from the same subject; performance is reported using AUC, balanced accuracy, macro-averaged F1-score, and confusion matrices for the three-class task, consistent with best practices for prediction model evaluation and reporting [24,25].

## 2.6. Robustness experiments

To assess robustness to acquisition variability (see Figure 5), three evaluation scenarios are defined: (i) within-protocol training and testing restricted to reference-protocol images, (ii) cross-protocol training on reference-protocol images with testing on low-dose and geometry-shift images, and (iii) multi-protocol training on a combined set of all three protocols with testing on each protocol subset separately. Performance degradation in the cross-protocol setting and its recovery in the multi-protocol setting serve as indicators of sensitivity to domain shift, which is a recognised challenge for imaging AI when acquisition conditions change across sites or time [22,23].

*2.7. Radiomic feature stability and importance*

For a subset of knees, multiple simulations with different noise realisations and small protocol perturbations are generated. Intraclass correlation coefficients (ICCs) are computed for each feature across repeats and protocols to quantify stability, and ICC form and reporting follow established guidance for reliability studies [38]. Feature stability is visualised in Figure 6 (left panel). Feature importance is assessed using model coefficients (logistic regression) and impurity-based scores (tree-based models). The most important features for KL-like "0" vs. "2" discrimination are shown in Figure 6 (right panel) and interpreted qualitatively in terms of joint space narrowing and subchondral texture.

## 3. Results

*3.1. Synthetic cohort characteristics*

In our work, the cohort comprises 260 knees distributed across KL-like grades as stated in Table 1. Each knee is imaged under three acquisition protocols, yielding 780 radiographs. Representative examples across grades and protocols are demonstrated in Figure 3.

Grade-dependent morphological patterns are clearly visible. KL-like grade "0" images show normal, symmetric joint spaces without osteophytes, grade "1" images display very subtle narrowing or tiny osteophytes, and grade "2" images demonstrate definite medial joint space narrowing, visible osteophytes, and slightly denser subchondral bone. Low-dose images have increased noise and slightly reduced contrast relative to the reference protocol, while geometry-shift images exhibit subtle changes in condyle projection and joint line orientation, mimicking clinical variability in positioning.

*3.2. Radiomic signatures and feature stability*

Radiomic analysis reveals consistent trends across OA severity. As KL-like grade increases, first-order mean intensity and energy increase, reflecting thicker, denser subchondral bone. Shape descriptors capture emerging varus/valgus deviation and marginal osteophyte development. GLCM contrast, entropy, and dissimilarity increase, while homogeneity de- creases, indicating coarser trabecular patterns. GLSZM and GLRLM features are sensitive to large homogeneous zones, and long runs grow with more extensive sclerosis.

The features stability across acquisition conditions are presented in Figure 6 (left panel). A subset of first-order and texture features shows high ICCs across reference, low-dose, geometry-shift, and test–retest simulations, forming a robust feature subset. Other features, particularly some higher-order textures, exhibit substantial variability, especially under low-dose conditions, indicating vulnerability to noise and protocol changes. The most important features for KL-like "0" vs. "2" discrimination are illustrated in Figure 6 (right panel). Many belong to the GLCM and GLSZM families and have interpretable associations with joint space narrowing or subchondral texture coarseness.

### 3.3. Classification performance

In the within-protocol setting (reference-protocol training and testing), binary classifiers distinguishing KL-like "0" from "2" achieve high AUC and balanced accuracy. Logistic regression with a small selected feature set performs strongly, while random forest and gradient boosting models provide comparable or slightly higher AUCs at the cost of reduced interpretability. Qualitative performance patterns across scenarios are demonstrated in Figure 5.

Three-class classification (KL-like "0" vs. "1" vs. "2") is more challenging, but yields promising macro-averaged F1-scores and confusion matrices with limited confusion between the extreme classes ("0" vs. "2"). Grade "1" behaves as an intermediate category, where overlap in radiomic signatures is expected biologically.

### 3.4. Robustness to acquisition variability

When models trained solely on reference-protocol images are evaluated on low-dose and geometry-shift images (cross-protocol scenario), performance decreases, highlighting sensitivity to domain shift, as demonstrated Figure 5 (middle panel). The magnitude of degradation depends on the model and on the contribution of unstable features to the predictive signature. Training on the combined multi-protocol dataset restores much of the lost performance when evaluated across each protocol subset, as illustrated Figure 5 (right panel). This indicates that synthetic multi-protocol training can pre-expose models to realistic acquisition variability and improve robustness, even before they encounter real-world imaging data.

## 4. Discussions

### 4.1. Principal findings

In this research, we introduce the physics-based synthetic simulation framework (PSSF) for generating synthetic knee radiograph cohorts and applied it to early OA radiomics research. By explicitly modelling knee morphology and the radiographic imaging chain, we create a virtual cohort spanning KL-like grades "0–2" and used it to design and stress-test radiomics and ML pipelines.

In this research, the main findings are: (i) synthetic radiographs encode early OA morphology in a way that radiomic features can detect and quantify, yielding plausible grade- dependent signatures, (ii) within the synthetic domain, radiomics-based classifiers achieve high discrimination between normal and mild OA and acceptable performance for three-class grading, (iii) radiomic feature stability varies substantially across features and acquisition conditions, and synthetic experiments help identify robust subsets (see Figure 6), and (iv) multi-protocol synthetic training improves robustness to acquisition domain shifts compared with single-protocol training (see Figure 5).

*4.2. Relation to prior work*

Deep learning for automated knee OA grading has shown strong promise, with models achieving high performance when trained on large, curated datasets of knee radiographs [10–12,14,15]. Such methods are data-hungry and can be affected by dataset shift when acquisition conditions or patient populations differ, motivating explicit robustness evaluation and external validation [15,23]. GAN-based synthetic knee radiographs have been proposed to alleviate data scarcity and privacy barriers. Prezja et al. demonstrated that "DeepFake" OA knee radiographs are sufficiently realistic to deceive medical experts and can augment classification models [30]. Follow-up work explored CycleGAN-based generation of hypo- thetical OA states and evaluated augmentation strategies in OA classification [28,31,32]. In contrast, our PSSF is purely physics-based and does not require access to real radiographs for training, offering transparent control over both morphology and acquisition parameters. Our focus on IBSI-compliant radiomics and feature stability aligns with broader standardisation efforts in radiomics [16–18]. Controlled synthetic cohorts can complement patient datasets by enabling systematic experiments on acquisition variability, feature repeatability, and harmonisation, which are well-recognised barriers to generalisable radiomic models [19–22]. Finally, this work is aligned with the wider discussion on safe deployment of AI in healthcare, where transparent reporting (TRIPOD, TRIPOD+AI, CLAIM) and stress- testing across plausible domain shifts are increasingly emphasised before real-world adoption [24–26,39].

*4.3. Role of physics-based synthetic cohorts*

Physics-based synthetic cohorts are not a replacement for clinical studies but can complement them in several ways. Firstly, they provide a methodological sandbox where radiomics pipelines, feature sets and model architectures can be prototyped and debugged before clinical deployment. Secondly, acquisition parameters, i.e., kVp, mAs, SDD, and angle, can be systematically varied to study their impact on radiomic feature stability and model performance, guiding protocol design without exposing patients. Thirdly, synthetic data is non-identifiable and can be shared as open benchmarks for multi-centre comparison of radiomics and AI methods. Finally, synthetic OA cases with known severity can be used for teaching radiographic interpretation and AI concepts to clinicians and trainees.

*4.4. Limitations*

This research has several limitations. The current simulator is 2D and focuses primarily on bony anatomy and joint space, with limited representation of soft tissues and complex pathologies (e.g., effusion, meniscal extrusion, or post-surgical change). Scatter and detector response are approximated rather than simulated with full Monte Carlo methods, so system-specific fidelity is limited; Monte Carlo toolkits (e.g., Geant4) and validated diagnostic X-ray beam simulation approaches could be used in future to increase physical realism and better match noise and scatter statistics [40,41]. The PSSF has not yet been calibrated against real radiographs or physical phantoms; external validation will be required to verify that radiomic feature distributions and image statistics align with clinical data. Only early OA (KL-like "0–2") is considered; modelling advanced disease would require additional anatomy and deformity representations. Finally, perfect labels and simplified anatomy may inflate apparent model performance compared with real-world settings, where observer variability in KL grading and cross-site acquisition differences are present [6,23].

*4.5. Future work*

Our future work can extend the PSSF to: (i) a 3D simulator with more detailed cartilage, meniscal and ligamentous structures, and supporting weight-bearing views, (ii) calibrate simulator parameters using phantoms or a small set of anonymised clinical images to better match intensity histograms and noise power spectra, (iii) combine physics-based simulation with data-driven generative models for enhanced realism and domain adaptation, and (iv) apply our framework to other joints and modalities, e.g., hip OA, CT, or MRI, to investigate generalizability of radiological imaging diagnoses.

## 5. Conclusions

Physics-based synthetic knee radiograph cohorts can provide controllable, label-perfect data for designing and stress-testing radiomics pipelines for early osteoarthritis (OA), without involving real patients or requiring ethical approval. Once calibrated to clinical systems, simulators can help optimize protocols, feature sets and model architectures before deployment, reducing dependence on large, annotated datasets in early-stage methodological work. In this paper, we introduce a physics-based synthetic simulation framework (PSSF) for generating synthetic knee radiograph cohorts and demonstrated its use in training and testing radiomics models for early OA. By providing controllable, label-perfect data with explicit acquisition variability. Such synthetic cohorts can support the design and optimization of robust radiomics pipelines, before they are confronted with real clinical datasets. This framework-based approach may be particularly valuable in the clinical settings with limited data access or ethical constraints and aligned with broader efforts, to deploy AI techniques in healthcare for a safe, transparent, and evidence-based manner [39].

Our key messages for this PSSF simulator are summarized as: (i) synthetic radiograph cohorts can be fully generated from physics-based simulation, avoiding patient data and ethical barriers in early-stage AI/radiomics development, (ii) radiomics signatures and machine learning (ML) models trained on synthetic knee radiographs can reproduce the expected patterns of early OA, including joint space narrowing and subchondral texture change, and (iii) training on multi-protocol synthetic data can improve the robustness to acquisition variability, concluding that synthetic cohorts can be used to pre-empt and mitigate domain shift before models are exposed to real-world medical imaging.

**Author Contributions**

**Funding**

This research received no external funding.

**Declaration of Competing Interests**

The authors declare no competing interests.

**Data Availability**

The original contributions presented in this research can be directly requested from the corresponding author (A.A-B.).